\pdfoutput=1
\documentclass[11pt]{article}

\usepackage[final]{acl}

\usepackage{times}
\usepackage{latexsym}
\usepackage{algorithm}
\usepackage{algpseudocode}
\usepackage{amsmath}
\usepackage{graphicx}
\usepackage{siunitx}
\usepackage{booktabs} 
\usepackage{multirow}
\usepackage{colortbl}
\usepackage{lscape}

\usepackage{microtype}
\usepackage{inconsolata}

\usepackage{placeins}
\usepackage[toc,page]{appendix}

\title{FLICK: Few-Label Intermediate Learning to Empower Low-Resource Languages for Web-Scale Language Diversity}

\author{
Ali Almutairi  \\
UNSW, Australia \\
\And
Abdullah Alsuhaibani \\
UTS, Australia \\
\And
Shoaib Jameel \\
University of Southampton \\
\AND
Aditya Joshi \\
UNSW, Australia \\
\And
Gelareh Mohammadi \\
UNSW, Australia \\
\And
Imran Razzak \\
MBZUAI. UAE \\
}

\begin{document}

\maketitle

\begin{abstract}
In low-resource (LR) language communities, web-based discourse suffers from a suboptimal availability of web technologies. This limits their population's ability to leverage language models to be applied in settings such as cybercrime prevention and cross-dialect communication. Therefore, it reinforces linguistic inequality compared to high-resource (HR) languages' privileges, which contradicts the 10th goal of the United Nations Sustainable Development Goals (SDGs). Consequently, training machine learning models with minimal supervision has gained significant research attention. While self-training methods have proven effective in semi-supervised learning, they remain vulnerable to errors accumulating from noisy pseudo labels (PLs). Moreover, most recent approaches to the few-label (FL) classification problem are either designed for HR languages or involve complex cascading models that are prone to overfitting. To address the persistent challenge of FL text classification in LR linguistic contexts, where existing methods often struggle with noisy PLs, model bias, limited accessibility, poor scalability, and high energy costs, we propose the FL Intermediate Learning technique to empower LR Languages for Web-Scale Language Diversity FLICK. FLICK introduces a novel PLs refinement component, a departure from traditional pseudo-labelling strategies by identifying and leveraging top-performing PLs clusters. This component specifically learns to distil highly reliable PLs from an initial broad set by focusing on single-cluster cohesion and leveraging an adaptive Top-K selection mechanism. This integrated architecture is crucial for mitigating the requirement for ample annotated data and high cost, allowing and empowering the participation of less represented populations on the web with less energy. We demonstrate FLICK's efficacy across 14 diverse datasets, encompassing challenging LR languages such as Arabic, Urdu, and Setswana, alongside English, showcasing its superior performance, scalability, and adaptability. We observe that introducing the K-aware PLs mechanism in the intermediate layer noticeably improves FL learning. FLICK leads to more dependable LR text classification systems that support fairer, more inclusive, and socially beneficial web applications.
\end{abstract}

\section{Introduction}
Lack of accessibility and scalability in data scarcity problems significantly limits the application of modern machine learning techniques, particularly in low-resource language settings, which foster inequality and non-inclusive indirectly. While there is sufficient data for some well-studied tasks in these languages, many tasks still suffer from poor data availability. Various deep learning algorithms have achieved notable success with high amounts of labelled data \cite{mironczuk2018recent}, typically numbering in the tens of thousands of examples. However, when labelled data are scarce, these complex neural networks tend to underperform \cite{labonte2024towards}. This has spurred interest in semi-supervised learning (SSL) approaches, which either augment existing labeled data or employ unlabeled data to alleviate these limitations, as unlabeled data is generally easier and more cost-effective to obtain.

SSL methods often rely on Neural Machine Translation (NMT) systems for data augmentation through back-translation \cite{rastgoo2021sign}. This process involves translating a sentence into another language and then back into the original language. However, this approach can be cumbersome in practice, as it requires an additional NMT system. Moreover, if the data distribution of the specific task differs from that used to pre-train the NMT system, the quality of the generated sentences may suffer. Furthermore, the phenomenon of `translationese' underscores the ways in which translated texts may occasionally lose cultural nuances \cite{intro-phenom-translat-Gellerstam1986TranslationeseIS}, leading to a linguistic representation that can appear less natural and culturally embedded compared to the original language. This divergence from the source material raises important considerations regarding the fidelity and authenticity of translated works in conveying the richness of cultural context. A recent study shows that such case affect the training model stage \cite{intro-translat-affect-train-yu-etal-2022-translate}.

An alternative approach is few-shot learning (FSL), which involves selecting \(k\) samples from each class (\textit{N-way-K-shot}) \cite{10.1145/3386252}, and is useful when labelled data is scarce \cite{LU2023109480}. Recent advancements in Natural Language Processing (NLP) have demonstrated that this challenge can be addressed by leveraging Large Language Models (LLMs) to provide a few examples during inference, thereby mitigating the issue of limited training data \cite{IDoFew}. Using this technique in inference models, particularly with BERT-based models or LLMs, presents certain challenges. The former is often limited by its smaller model size and parameters, while the latter, despite its greater knowledge, incurs high computational costs. Furthermore, both types of models may struggle with LR languages, as they have primarily been developed based on HR languages.

Nevertheless, fine-tuning Bidirectional Encoder Representations from Transformers (BERT), such as BERT model \cite{devlin2019bertpretrainingdeepbidirectional} or AraBERT \cite{antoun-etal-2020-arabert} (a language representation model for Arabic, based on Google's BERT), helps address the limited label challenge, but its performance remains constrained \cite{goyal2023finetune}. An alternative approach is self-training, where PLs are generated for unlabeled data, which are then treated as labelled data for subsequent training. Traditional self-training methods typically do not include sample selection or account for noise in the PLs. To address this, Shnarch et al. propose $Cluster\&Tune$ \cite{shnarch-etal-2022-cluster}, introducing an intermediary task aimed at enhancing the fine-tuning performance of BERT. Their approach involves leveraging unsupervised learning, which generates PLs that are subsequently utilised as inputs for fine-tuning. Another notable contribution in this area is IDoFew \cite{IDoFew}, which incorporates dual clustering, with the second cluster serving as a correction mechanism. However, despite the elegance and efficacy of these methodologies, their applicability to LR languages remains largely unexplored. In particular, the effectiveness of fine-tuning with a limited number of labelled examples, especially in the context of the Arabic language, remains an open question.

\begin{figure*}[!]               
    \centering
    \includegraphics[
        trim={2cm 1cm 0 0},clip,
        width=\linewidth]{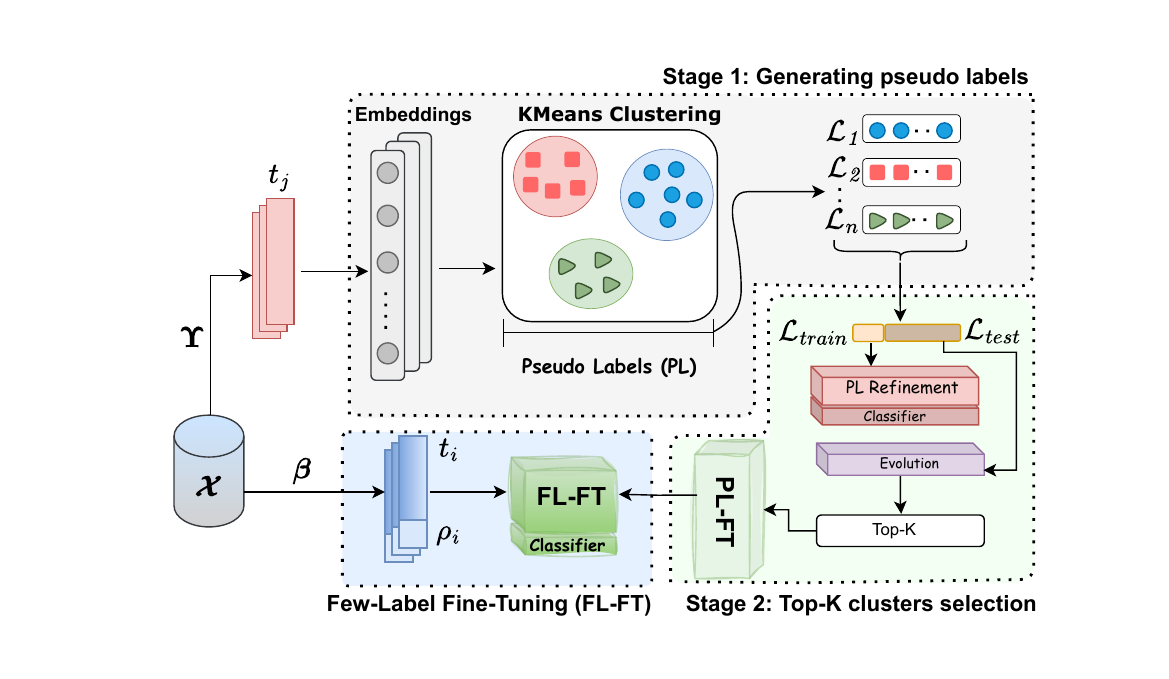}

\captionof{figure}{Our novel \textsc{FLICK} framework.  
}
\label{fig:FLICK-methodology-diagram}
\end{figure*}

To address the challenges associated with fine-tuning BERT-based models in LR languages with limited labelled data, particularly in the presence of large unlabeled datasets, we introduce FLICK, a novel framework tailored for FL text classification, specifically in languages like Arabic. Inspired by the inter-training mechanism employed in \cite{IDoFew}, our approach leverages clustering algorithms to generate an initial set of PLs (e.g., K-means with k=20 clusters, as detailed in Algorithm 1) from the unclassified data, as shown in Figure \ref{fig:FLICK-methodology-diagram}. Recognising that not all initial pseudo-label clusters are equally reliable, a critical secondary stage is implemented: the Pseudo-Label (PL) Refinement component. This component is essential to identify and select the most pertinent and higher quality PLs clusters (the Top-K clusters, as per Algorithm 2) to mitigate the propagation of noisy or ambiguous labels. Subsequently, these refined PLs are used to train an intermediate a Pseudo-Label Fine-Tuning (PL-FT) model, whose learned knowledge is then leveraged to fine-tune a Few-Label Fine-Tuning (FL-FT) model with the limited set of real labels.

Our key contributions are as follows: \textbf{(1)} We develop \texttt{FLICK}, a fundamentally new framework for FL text classification designed specifically to overcome the PLs noise inherent in LR linguistic environments. Our core contribution lies in the PLs refinement component, which represents a significant algorithmic advancement over existing semi-supervised and few-shot methods. \textbf{(2)} Implementation of a Top-K selection method among the clusters to ensure the quality of data points for fine-tuning. \textbf{(3)} Our extensive experiments across 14 diverse datasets, including severely LR languages such as Arabic, Urdu, and Setswana, empirically demonstrate that FLICK consistently outperforms state-of-the-art (SOTA) FL baselines, including direct comparisons with IDoFew. This validates the effectiveness of our PLs refinement component in mitigating PLs noise and achieving robust classification performance where traditional multi-cluster approaches falter due to data scarcity and linguistic peculiarities.

\section{Related Work}
Despite its ubiquity, web data often presents itself as unlabelled, or labeled to a small extent. To address some of these issues, various methods have been developed, involving the use of a small number of labels from diverse perspectives relevant to the specific task \cite{IDoFew}. This is known as the `FL' setting, and implies that a small portion of the dataset is labelled for a given task \cite{wang2020generalizing-Lit-P1}, such as text classification \cite{liang2024selp-Lit-P1}. The concept of leveraging the ``few labels'' method has proven advantageous in addressing different issues such as cyberbullying, cybercrimes, hate speech, spam, phishing, and offensive content detection 
, primarily due to the scarcity of annotated labels in these domains and the fast-evolving nature of the issues themselves. Consequently, the FL techniques have gained widespread adoption across various domains and data types, encompassing both image and text data.

Researchers have addressed the scarcity of fully annotated datasets challenge through various methods, including synthetic data generation \cite{Lit-P2-augmen-ijcai2023p565}, semi-supervised learning \cite{Lit-P2-semi-lee-etal-2024-superst}, transfer learning \cite{Lit-P2-tran}, and few-shot learning \cite{Lit-P2-few}, each of which has shown promising results. However, employing these methods individually has revealed inherent limitations, thus creating ample space for improved solutions.

Although synthetic data methods can produce PLs, they may introduce artificial noise. Additionally, semi-supervised learning relies on the entire set of generated PLs, potentially leading to inefficient training by incorporating classes with limited observations, thus compromising the model's effectiveness. Furthermore, the main challenge with transfer learning in FL low-resource settings is that fine-tuning foundational models on extremely limited labelled data often leads to overfitting and poor generalisation. Lastly, FSL may not leverage the latent knowledge in unlabeled data and inherit the limitations posed by BERT-based models and LLMs. 

Recently, Intermediate Learning (IL) has emerged as a promising methodology for addressing the challenges of FL tasks. IL offers various approaches to tackle this issue. \citet{shnarch-etal-2022-cluster} introduced the Cluster and Tune framework, which has been recognised as SOTA in the scientific community. Additionally, \citet{IDoFew} developed an optimised version of the previous architecture called IDoFew, which surpassed the results of their predecessor on the same benchmarks. However, their use of multiple clusters slowed down the model and incurred additional costs. 

Many of the above techniques are primarily used for HR languages, although some have been adapted for LR languages such as Arabic and its widely spoken dialects. \cite{alyami-al-zaidy-2022-weakly} employed a semi-supervised technique to tackle the limited labelled data issue within the context of Arabic dialects, utilising two datasets: (1) Arabic Dialect Short Text and (2) Arabic Dialect Dictionary. They obtained significant results by incorporating the AraRoBERTa model. Additionally, \cite{AMEUR20233898-lit-p4} demonstrated that a semi-supervised approach achieved 91\% accuracy in sarcasm detection using a limited labelled dataset. However, accuracy alone may not suffice as an evaluation metric, especially in cases of dataset imbalance. Furthermore, IL could potentially enhance the current state, particularly in LR languages like Arabic. 

Our approach markedly diverges from the aforementioned techniques. Cluster\&Tune \cite{shnarch-etal-2022-cluster} utilised four distinct methods in the inter-training phase prior to the fine-tuning of the language model. Despite their notable achievements, the deployment of multiple BERT models rendered the process costly. In response, the IDoFew framework halved the number of models used during inter-training and introduced a fraction of training instances in the subsequent stage, thus boosting performance and contributing to SOTA outcomes. Nevertheless, their inter-training still involved complex methods such as the use of several clusters and was limited to English data. Our method, however, is designed not only for effectiveness but also for speed, potentially outpacing the IDoFew model. Moreover, we have refined the concept of fraction random sampling by selecting the top pseudo-labelled samples for ongoing learning.

Furthermore, compared to previous studies, we focused on LR languages such as Arabic, which are more difficult to model than English. Some challenges include script complexity, morphological complexity, ambiguity and homography, data scarcity and quality, and dialect variation. We enhance the architecture by improving clusters by integrating a PLs Refinement component, functioning as a BERT enhancer, implemented through the utilisation of language-related BERT models such as the AraBERTv2 model to select higher quality PLs. Unlike semi-supervised labelling, we select only the top PLs instead of the entire set. Furthermore, by training the top PLs in the PL-FT step on the base BERT, such as AraBERTv2 and transferring this knowledge to the FL-FT model, we eliminate the assumptions of traditional transfer learning and enhance the original BERT's knowledge for the specific task at hand.

\section{Our Novel FLICK Framework}
We tackle the issue of learning with FL, especially in LR languages such as Arabic, in a text classification setting. In doing so, we not only confront the difficult task of modelling with a limited number of labels but also take into account datasets with scarce resources. This presents a greater challenge compared to previous works such as \cite{IDoFew}, given the nature of the dataset. For example, employing a machine learning to model the Arabic language introduces unique challenges due to its complex linguistic structure and cultural subtleties. Arabic is a richly inflected language, where words acquire new meanings by adding prefixes, suffixes, and internal modifications. Arabic also exhibits significant diversity in terms of the dialects of Arabic. FL learning necessitates a minimal amount of labelled data, which poses a challenge for Arabic, given the scarce annotated datasets. The limited data may also suffer from an imbalanced label distribution, with certain classes having far fewer examples than others, impeding the model's learning effectiveness. Additionally, the multitude of Arabic dialects adds complexity to FL learning, as the model must be able to generalise to unfamiliar dialects with scant training data.

In our setting, the dataset for training a model contains significantly fewer labelled examples than ideal. This usually can lead to suboptimal performance due to insufficient information to learn patterns and relationships. Our \texttt{FLICK} framework addresses this by employing a novel two-stage approach: (1) PLs generation and initial Clustering (Stage 1), where we harness unlabeled data to create initial semantic groupings; and (2) PLs refinement and intermediate fine-tuning (Stage 2), where higher quality PLs are distilled and used to train a powerful intermediate classifier. This intermediate learning phase is crucial for enhancing the model's generalisability and enriching the base knowledge of the fine-tuned model before exposure to a limited set of real labels. We have depicted the high-level overview of our model in Figure~\ref{fig:FLICK-methodology-diagram}.


As illustrated in the Figure~\ref{fig:FLICK-methodology-diagram}, the input is a text dataset whose semantic representations are obtained using a language model. Specifically, each sentence is encoded into a low-dimensional embedding representation that captures its semantic content. These representations are then passed into a K-means clustering model that semantically groups them into distinct clusters, from which PLs are generated. These PLs are then used to fine-tune an existing language-specific pre-trained classifier, PL-FT. The classifier is designed to predict the correct PLs for new, unseen data. The classifier's performance is continuously evaluated and improved. This involves adjusting the model's parameters to enhance its accuracy. Initially, K-means clustering (with (k=20), as detailed in Algorithm 1, groups these embeddings into distinct pseudo-labelled clusters. \texttt{FLICK} then employs a novel PLs refinement component (detailed in Algorithm 2) where the classifier's performance is evaluated per cluster. Based on this evaluation, the model's predictions are refined by training the model on the samples from the Top-K clusters. This approach is particularly useful for LR languages like Arabic, where labelled data is scarce. Finally, the PL-FT learnt weights are then used to initialise FL-FT, which will be fine-tuned on the real labels. 

The mathematical formulation is as follows. Given a dataset \(\mathcal{X}\) defined by two subsets: \(\beta\) and \(\Upsilon\). The subset \(\beta\) is a supervised set \(\{ (t_i, \rho_i) \}_{i=1}^\varepsilon\), where \(t_i = \{t_1, t_2, t_3, \ldots, t_\varepsilon\}\) represents the text, and \(\rho_i \in \{1, 2, 3, \ldots, Z\}\) denotes the corresponding labels for each \(t_i\). Each text \(t_i\) is described by \(\lambda\)-dimensional vectors. The labels \(\rho_i\) can represent either binary or multiclass classification. The subset \(\Upsilon\) represents the unlabeled data, denoted by \(\{t_j\}_{j=1}^{\nu}\), containing text instances with no labels. It is worth mentioning \(\varepsilon \) is much less than \(\nu\). Our objective is to harness the potential of unlabeled datasets through the implementation of the intermediate method, resulting in a viable solution for LR languages and FL fine-tuning challenges.

\subsection{FLICK's Architecture} 
FLICK employs two distinct clustering algorithms to generate PLs for each cluster. This approach minimizes the risk of inadequate pseudo-labelling, which is particularly pertinent for Arabic texts where semantic content can be highly ambiguous. 

FLICK involves two key training stages: (1) PLs Generation (Algorithm 1): where the unlabeled dataset is clustered (e.g., into 20 initial clusters) to generate PLs; (2) PLs Refinement and Intermediate Fine-Tuning, PL-FT: Given that noisy PLs or an excessive number of unreliable clusters can compromise performance, this stage focuses on identifying and leveraging the top-performing clusters as a highly representative subset of the dataset using a language-specific BERT model. This process is detailed in Algorithm 2 and enables the intermediate classifier PL-FT to acquire robust representations.

\begin{algorithm}[t]
\scriptsize
\caption{Generate Pseudo Labels} \label{alg:generate-pseudo-labels}
\begin{algorithmic}[1]
\State \textbf{Input:} Unlabeled data
\State \textbf{Output:} Pseudo labels
\State \textbf{Load:} Sentence Transformer model 
\State \textbf{Convert unlabeled data into dense vectors:}
\State \hspace{\algorithmicindent} Let \(\Upsilon\) unlabeled set. For each text \(\{t_j\}_{j=1}^{\nu}\), convert it into a token sequence $\hat{t_j}$ = $\{tok_1, tok_2, \ldots, tok_\nu\}$.
\State \hspace{\algorithmicindent} Feed $\hat{t_j}$ into the Sentence Transformer which computes the embeddings:
\State \hspace{\algorithmicindent} $\mathbf{\lambda}_j = \frac{1}{m} \sum_{j=1}^m \text{BERT}_{\text{LM}}(\hat{t_j})$, where $\text{BERT}_{\text{LM}}(\hat{t_j})$ is the embedding of token.
\State \textbf{Apply:} K-Means algorithm with $k = 20$ on the \(\lambda\) set of vectors $\{\mathbf{\lambda}_1, \mathbf{\lambda}_2, \ldots, \mathbf{\lambda}_n\}$:
\State \hspace{\algorithmicindent} Initialize $k$ centroids randomly. 
\State \hspace{\algorithmicindent} Repeat until convergence:
\State \hspace{\algorithmicindent}\hspace{\algorithmicindent} Assign each vector $\mathbf{\lambda}_i$ to the nearest centroid, forming clusters.
\State \hspace{\algorithmicindent}\hspace{\algorithmicindent} Update each centroid to be the mean of the vectors assigned to it.
\State \textbf{Assign:} cluster labels as pseudo labels to the data.
\end{algorithmic}
\end{algorithm}

Our training protocols consist of two key stages. Stage 1 is described in Algorithm \autoref{alg:generate-pseudo-labels}, and Stage 2 is described in Algorithm \autoref{alg:top-k-extr}.

\noindent\textbf{Stage 1: Generating pseudo labels:} Our goal is to generate distinct clusters of PLs with consistent patterns from unlabelled data using embedding and k-means clustering algorithms, which yield more reliable PLs for subsequent analysis. The unlabelled data is preprocessed to remove stop words and punctuation, and apply stemming before being fed into the BERT-based model known as paraphrase-multilingual-MiniLM-L12-v2 for vector representation. The paraphrase-multilingual-MiniLM-L12-v2 model, accessed via the sentence-transformers library \cite{Flick-methodology-sentence-transformers}, is capable of producing embeddings for sentences in multiple languages, resulting in dense vector representations. Notably, this embedding approach has proven to offer a more accurate representation of the Arabic language than previous methods centred on English, due to its effectiveness in capturing semantic meanings. These embeddings are then used as input for the k-means clustering algorithm.

In the k-means algorithm, when the $k$ centroids converge to the nearest point within the clusters, PLs are assigned to each cluster. We have refined this process to enhance the representation of PLs in FLICK. This is achieved by selecting the Top-K performing clusters and discarding the rest, which results in more accurate PLs and the removal of noisy clusters.

\begin{algorithm}[t]
\scriptsize
\caption{Top-K Extraction and Selection} \label{alg:top-k-extr}
\begin{algorithmic}[1]
\State \textbf{Input:} Pseudo labels clusters
\State \textbf{Output:} Top pseudo label clusters
\State \textbf{Divide:} The input into training and test datasets:
\State \hspace{\algorithmicindent} Let $\mathcal{L}$ the set of data points with pseudo labels. 
\State \hspace{\algorithmicindent} Training set $\mathcal{L_{\text{train}}}$ = \{$\iota_1$, \ldots, $\iota_{0.25n}$$\}$  (25\% of the data)
\State \hspace{\algorithmicindent} Testing set $\mathcal{L_{\text{test}}}$ = \{$\iota_{0.25n+1}$, \ldots, $\iota_n$$\}$ (75\% of the data)
\State \textbf{Feed:} $\mathcal{L_{\text{train}}}$ into the language-related BERT model for initial training:
\State \hspace{\algorithmicindent} Optimize the model parameters $\theta$ by minimizing the loss function  $\ell$,  over $\mathcal{L_{\text{train}}}$
\State \textbf{Feed:} $\mathcal{L_{\text{test}}}$ into the trained language-related BERT model to refine and evaluate:
\State \hspace{\algorithmicindent} Compute predictions $\hat{y_i}$ = $\text{BERT}_{\theta}$($\iota_n$) for each $\iota_n$ in $\mathcal{L_{\text{test}}}$
\State \hspace{\algorithmicindent} Evaluate model performance based on metrics such as accuracy and loss
\State \textbf{Identify:} top clusters from $\mathcal{L_{\text{test}}}$:
\State \hspace{\algorithmicindent} Rank clusters based on correctly predicted culsters
\State \hspace{\algorithmicindent} Select top performing clusters, denoted as Top-K
\State \textbf{Return:} Top-K as the final output
\end{algorithmic}
\end{algorithm}

\noindent\textbf{Stage 2: Top-K clusters extraction and selection:} This stage, encompassing the core intermediate learning of \texttt{FLICK}, refines the initial PLs generated in Stage 1 and trains an intermediate classifier. The process involves two key steps:

\begin{itemize}
    \item From the complete set of pseudo-labelled data (generated in Stage 1), we first divide it into an internal training set $\mathcal{L_{\text{train}}}$ and an internal testing set $\mathcal{L_{\text{test}}}$. A vanilla language BERT model (e.g., AraBERTv2 for Arabic) is initially fine-tuned on $\mathcal{L_{\text{train}}}$. This model then evaluates $\mathcal{L_{\text{test}}}$, and the clusters are ranked based on the Top-K correctly predicted clusters on $\mathcal{L_{\text{test}}}$. The Top-K performing clusters, denoted as Top-K, are then identified as the most reliable subset of pseudo-labelled data.

    \item The selected Top-K pseudo-labelled clusters are then used to further fine-tune a language-specific BERT model (e.g., AraBERTv2 for Arabic) to create our PL-FT model. This training step may solidify the model's understanding of the semantic space represented by higher quality PLs, enhancing its generalisability. This is the intermediate learning phase where the model significantly augments its knowledge base. The weights learned by the PL-FT model serve as crucial initialisation for the final FL fine-tuning.
\end{itemize}


\noindent \textbf{Few-Label Fine-Tuning (FL-FT):} The knowledge acquired during the earlier learning PL-FT phase has allowed for effectively utilising a small set of labelled data in resource-constrained environments. We employed multiple datasets from various domains and fine-tuned the model separately on each dataset for their corresponding downstream tasks. Finally, the weights obtained from the previous stage (PL-FT model) are used in FL-FT model to further fine-tune on a few true labels.

\section{Experiments and Results}
\begin{table}[t]
\centering
\scriptsize
\setlength{\tabcolsep}{3pt}
\renewcommand{\arraystretch}{0.82}

\begin{tabular*}{\linewidth}{@{\extracolsep{\fill}}l S[table-format=6.0] S[table-format=5.0] S[table-format=2.0] l l}

\toprule
\textbf{Dataset}
& {\textbf{Train}}
& {\textbf{Test}}
& {\textbf{$C$}}
& \textbf{Task}
& \textbf{Lang.} \\
\midrule
MPOLD        &   3400 &   600 &  2 & Offensive & Arabic   \\
DART         &    972 &   455 &  5 & Dialects  & Arabic   \\
Sa'7r        &   3959 &  1187 &  2 & Sarcasm   & Arabic   \\
ArBNTopic    &  15827 &  1583 & 14 & Domain    & Arabic   \\
Sentiment-Ar &   8437 &  2110 &  3 & Sentiment & Arabic   \\
Sarcasm-Ar   &   8437 &  2110 &  2 & Sarcasm   & Arabic   \\
Dialects-Ar  &   8437 &  2110 &  5 & Dialects  & Arabic   \\
RUD          & 110385 & 36795 &  2 & Offensive & Urdu     \\
RUED         &   2306 &   769 &  4 & Sentiment & Urdu     \\
Setswana     &   3893 &   974 & 10 & Domain    & Setswana \\
\bottomrule
\end{tabular*}

\caption{Datasets with full train/test sizes, class count $C$, task type, and language.}
\label{tab:train-test-sizes}

\end{table}

\subsection{Datasets}


We evaluated the FLICK framework on seven different publicly accessible Arabic datasets, which include (1) ArBNTopic \cite{albared-etal-2023-arabic-Exp-dat-p1} (2) Sentiment \cite{abu-farha-magdy-2020-arabic-Exp-dat-p1} (3) Sarcasm \cite{abu-farha-magdy-2020-arabic-Exp-dat-p1} (4) The Saudi Dialect Irony (Sa'7r) \cite{almazrua-etal-2022-sa7r-Exp-dat-p1} (5) Dialectal Arabic Tweets (DART) \cite{alsarsour-etal-2018-dart-Exp-dat-p1} (6) Dialects \cite{abu-farha-magdy-2020-arabic-Exp-dat-p1} (7) Multi Platforms Offensive Language Dataset (MPOLD) \cite{chowdhury-etal-2020-multi-Exp-dat-p1}. The Sentiment, Sarcasm, and Dialects datasets were derived from the ArSarcasm dataset. Additionally, we extend the \texttt{FLICK} framework assessment on Urdu datasets: Roman Urdu Dataset (RUD) \cite{exp-data-p1-rud} and RUED \cite{exp-data-p1-rued} and one of the Bantu languages: Setswana \cite{exp-data-p1-setswana}. 

As a case study, we have also tested the performance on four English datasets (1) SMS Spam \cite{Exp-Dat-p1-sapm}, (2) Polarity\cite{Exp-Dat-p1-plor-pang-lee-2005-seeing}, (3) Yahoo \cite{Exp-Dat-p1-restEng-NIPS2015_250cf8b5}, and (4) DBpedia \cite{Exp-Dat-p1-restEng-NIPS2015_250cf8b5}. 

We have standardised the datasets by focusing on the ``text'' and ``label'' columns in the publicly shared datasets. Considering our datasets must contain only a limited number of labels, we have manually reduced the number of labelled instances to simulate the FL learning problem; for example, we retained only 100 labelled instances. For more details on test/training size, please refer to Table~\ref{tab:train-test-sizes}.


\begin{table*}[t]
\centering
\scriptsize
\resizebox{\textwidth}{!}{
\begin{tabular}{lcccccccccccccc}
\hline \hline
\textbf{Models\textbackslash Datasets} & \multicolumn{2}{c}{\textbf{ArBNTopic}} & \multicolumn{2}{c}{\textbf{Sentiment}} & \multicolumn{2}{c}{\textbf{Sarcasm}} & \multicolumn{2}{c}{\textbf{Sa'7r}} & \multicolumn{2}{c}{\textbf{DART}} & \multicolumn{2}{c}{\textbf{Dialects}} & \multicolumn{2}{c}{\textbf{MPOLD}} \\ 
\cmidrule(lr){2-3} \cmidrule(lr){4-5} \cmidrule(lr){6-7} \cmidrule(lr){8-9} \cmidrule(lr){10-11} \cmidrule(lr){12-13} \cmidrule(lr){14-15}
 & \textbf{ACC} & \textbf{F1} & \textbf{ACC} & \textbf{F1} & \textbf{ACC} & \textbf{F1} & \textbf{ACC} & \textbf{F1} & \textbf{ACC} & \textbf{F1} & \textbf{ACC} & \textbf{F1} & \textbf{ACC} & \textbf{F1} \\ \hline
\multicolumn{15}{c}{\textbf{Baseline Models }} \\

\textbf{ArabicBERT} & 36.31 & 27.76 & 61.56 & 50.75 & 80.90 & 53.56 & 58.72&	57.53&	69.23&	61.68 &
       71.42 & 25.20 & 84.00 &	68.34
 \\

\textbf{AraBERTv2} & 51.68 & 38.63 & 66.30 & 57.62 & 80.71 & 61.98 &61.08 & 61.08 & 72.53 & 63.98& 74.79 & 28.15 & 83.83 & 66.92 \\
	
\textbf{AraBERT-twitter} & 42.29 & 32.50 & 65.26 & 53.89 & 83.65 & 55.97 & 60.99&	60.58&	73.85 &	66.31&
       74.03 & 26.84 & 86.33 & 68.76\\
\textbf{AlcLaMv2} & 12.30 & 4.47 & 52.84 & 26.95 & 83.65 & 45.55 & 47.09 &	45.16 &	45.93 &	35.93 & 66.82 & 16.02 & 83.33 & 45.45 \\
\textbf{CamelBERT-mix} & 38.25 & 27.53 & 64.03 & 52.08 & 83.65 & 46.95 & 56.78 &	55.39  & 72.31 & 63.50 & 72.18 & 24.26 & 85.83 & 72.09  \\
\textbf{CamelBERT-MSA-mix} & 36.48 & 29.48 & 62.65 & 50.72 & 83.41 & 48.66 & 61.67  & 61.64 &	69.01 &	60.79 & 73.22 & 26.52 & 81.33 & 60.04
  \\
\textbf{MARBERTv2} & 24.22 & 17.99 & 63.70 & 46.05 & 83.65 & 45.55 & 59.39 &	58.95  &	74.50  &	66.09
    & 66.82 & 16.03 & 86.83 & 67.45 \\
\textbf{BERT} & 10.66 & 4.13 & 58.67 & 41.09 & 83.65 & 45.55 & 56.44 &	56.40&	32.75 & 	28.23 
    & 66.82 & 16.02 & 83.33 & 45.45 \\
\textbf{RoBERTa} & 9.06 & 1.19 & 57.30 & 37.83 & 83.65 & 45.55 &48.53&	47.37&	21.54&	11.53
      & 66.82 & 16.02 & 83.33 & 45.45 \\
\textbf{DistilBERT} & 9.90 & 2.49 & 57.63 & 38.61 & 83.65 & 45.55 & 47.01 &	44.59&	25.49 &	18.93
     & 66.82 & 16.02 & 83.33 & 45.45 \\

\hline
\multicolumn{15}{c}{\textbf{Fine-tuned LLMs via QLoRA }} \\
\textbf{Llama3 8B} & 10.39  & 09.59 & 36.58 & 33.32 & 68.33 & 71.61 & 58.38 & 62.12 & 19.14 & 20.51 & 44.92 & \textbf{51.75} & 69.17 & 72.29 \\
\textbf{AceGPT 16B} & 09.30 & 07.47 & 30.19 & 34.85 & 81.53 & \textbf{78.92} & 59.30 & \textbf{62.72} & 19.56 & 25.72 & 38.45 & 46.36 & 71.33 & 71.69 \\
\hline

\multicolumn{15}{c}{\textbf{State-of-art Models}} \\
\textbf{Cluster\&Tune} & 53.33 & 47.03 & 51.00 & 28.76 & 83.65 & 45.55 & 52.90 & 52.83 & 22.63 & 16.20 & 66.82 & 16.02 & 83.33 & 45.45 \\
\textbf{Cluster\&Tune{$_{AraBERTv2}$}} & 53.33 & 47.03 & 65.69 & 56.25 & 82.46 & 56.19 & 58.38 & 57.64 & 69.01 & 60.56 & \textbf{75.36} & 28.11 & 81.66 & 65.60 \\ 

\textbf{IDoFew} & 17.52 & 10.96 & 58.06 & 40.42 & 83.65 & 45.55 & 59.98 & 59.41 & 40.00 & 32.90 & 66.82 & 16.02 & 83.33 & 45.45 \\
\textbf{IDoFew{$_{AraBERTv2}$}}+ & 59.27 & 49.53 & 64.22 & 59.23 & 81.18 & 60.49 & 61.75 & 61.03 & 67.91 & 59.72 & 72.84 & 28.31 & 82.83 & 64.87 \\
\textbf{IDoFew{$_{AraBERTv2}$}}* & 59.27 & 49.53 & 64.12 & 59.05 & 81.18 & 60.49 & 
 58.88 & 56.79 & 71.86 & 62.78 & 72.80 & 28.28 & 84.66 & 64.95 \\ \hline
\multicolumn{15}{c}{\textbf{Our Model}} \\

\textbf{FLICK } & \textbf{63.48} & \textbf{53.58} & \textbf{66.45} &\textbf{ 61.05} & \textbf{84.31} & 62.25 & \textbf{62.85} & 62.24 & \textbf{75.60} & \textbf{67.12} & 75.21 & 29.08 & \textbf{87.50
} & \textbf{72.81} \\ \hline
\end{tabular}
}
\caption{Performance comparison across different models on Arabic datasets. The all-MiniLM-L6-v2 embedding method is denoted as a + symbol whereas the paraphrase-multilingual-MiniLM-L12-v2 embedding method is denoted as a * symbol. 
}
\label{table:performance_comparison}
\end{table*}

\subsection{Comparative Models}

\emph{\textbf{BERT models:}} We have compared FLICK with popular publicly available language models that include BERT \cite{BERT-Exp-compMod-p1-devlin-etal-2019-bert}, RoBERTa \cite{Roberta-Exp-compMod-p1-liu2019robertarobustlyoptimizedbert}, DistilBERT \cite{DistilBERT-Exp-compMod-p1-sanh2020distilbertdistilledversionbert}, and BantuBERTa \cite{Bantu_embeddings1}, all of which are built on the transformer architecture.

\noindent \emph{\textbf{Arabic-BERT models:}} We have also used the ArabicBERT in our study. These models are: ArabicBERT \cite{ArabicBERT-Exp-compMod-p1-etal-2020-kuisail}, AraBERTv2 \cite{AraBERTv2-Exp-compMod-p1-antoun-etal-2020-arabert}, AraBERT-Twitter \cite{AraBERTv2-Exp-compMod-p1-antoun-etal-2020-arabert}, AlcLaMv2 \cite{alclam-Exp-compMod-p1-ahmed-etal-2024-alclam}, CamelBERT-Mix \cite{camelbert-mix-Exp-compMod-p1-inoue-etal-2021-interplay}, CamelBERT-MSA-Mix \cite{camelbert-mix-Exp-compMod-p1-inoue-etal-2021-interplay}, and MARBERTv2 \cite{MARBERTv2-Exp-compMod-p1-abdul-mageed-etal-2021-arbert}. 

\noindent \emph{\textbf{LLMs:}} We compare FLICK with fine-tuned LLMs via QLoRA, given their demonstrated proficiency in handling complex tasks. For this purpose, we select Llama3-8B \cite{Res-Comp-p2-Llama-3} and AceGPT \cite{Res-Comp-p2-AceGPT-zhu2024second} as representative models in our comparison. 


\noindent \emph{\textbf{Intermediate models:}} To validate the effectiveness of our intermediate stage, we compared FLICK performance with Cluster \& Tune \cite{shnarch-etal-2022-cluster} and IDoFew \cite{IDoFew}, which both make use of an inter-training task.  We also substitute BERT with language-specific BERT such as AraBERTv2, to ensure a fair comparison for Arabic datasets, and a similar approach for the other languages.

We utilised the Adam algorithm \cite{kingma2017adammethodstochasticoptimization} as an optimiser with a learning rate ($\iota$) of 3e-5. The epoch and batch size were set to 10 and 64, respectively. This setup was employed for the BERT-based model in the final stage. As for Intermediate models, we used the parameters recommended in the original work.

\subsection{FLICK Settings}
Our experimental approach involves two consecutive intermediate training stages before the fine-tuning process with a few labelled samples. In the first stage, we convert all unlabeled data to embeddings using the sentence transformer and apply the paraphrase-multilingual-MiniLM-L12-v2 model, except for the Dialects dataset, where we use all-MiniLM-L6-v2. This process yields dense embedding vectors for input to a k-means cluster algorithm. Here, k is set to 20, and the maximum iteration is capped at 300.

Moving to the second stage, we initially divide the pseudo-labelled clusters into training and testing sets, with the split being 0.25 for training and 0.75 for testing. Furthermore, the split is stratified to ensure balanced partitioning between classes and mitigate bias, avoiding the issue of a training set containing only one cluster due to its small portion. Subsequently, the training set is passed to the AraBERTv2 with the following parameters: (1) Batch size = 64, (2) learning rate (LR) = 3e-5, (3) Epsilon = 1e-06, (4) number of epochs = 10, and (5) Adam Optimiser. After completing the training, we evaluate the model using the test set and identify the top 15 clusters from the experiment. Finally, we train the top 15 pseudo-labelled clusters in the AraBERTv2 model on the samples from the top 15 pseudo-labelled clusters, maintaining the same parameters as previously mentioned, except for the epoch set to 1 for the inter-training stage. For the fine-tuning of the main task, we train a small number of labelled samples, specifically 100 labelled data points. The training model is initialised with the weights from the last training process in the intermediate learning stage.

\begin{figure}[ht!]
    \centering
    \scriptsize
    \includegraphics[width=0.50781\textwidth]{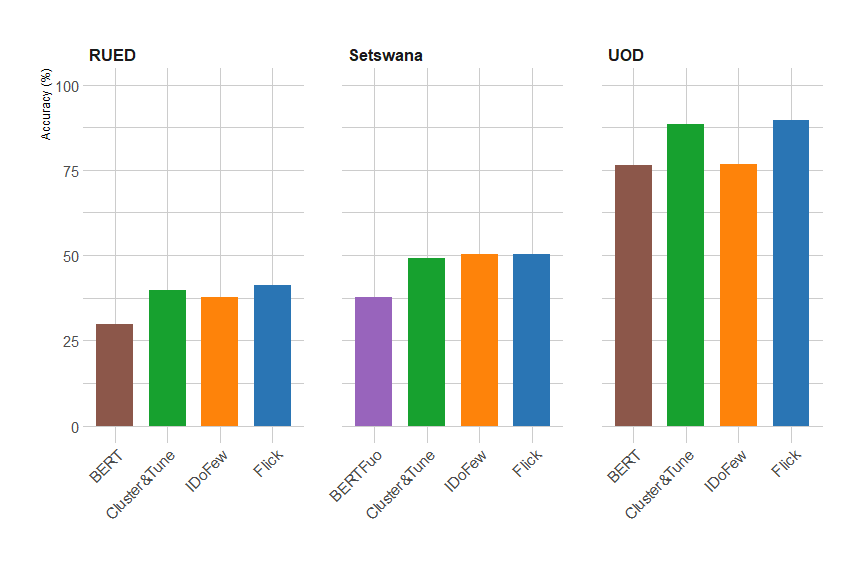}
    \centering
    \caption{Comparative analysis of the FLICK model alongside baseline and state-of-the-art models across various Urdu datasets (UOD and RUED) and Setswana dataset. The analysis is structured into three individual bar charts, each dedicated to a specific dataset.}
    \label{fig:low-resource-langs}
\end{figure}

\subsection{Results and Analysis}
We employed intermediate learning by integrating clustering and BERT-based algorithms in the learning phase using unlabeled data. This was followed by a fine-tuning stage to tackle the FL text classification problem in LR scenarios. We validated our framework on seven Arabic datasets, two Urdu, one Setswana, and 4 English datasets from various domains, and for most datasets, our model achieved SOTA results for FL text classification in LR contexts across different domains. The results of our experiments are summarised in Table \ref{table:performance_comparison}, where we compared the performance of our model with BERT-based models, LLMs and previous related work.

In the ArBNTopic dataset, \texttt{FLICK} achieved an accuracy of 63\% and an F1-score of 53\%, outperforming the BERT-based models and LLMs baseline, Cluster \& Tune, and IDoFew models. The difference between our model and the closest baseline model was 12\%. However, the IDoFew model reduced this difference to 4\% when the same model and embedding technique used in our framework were applied to the IDoFew original work. This reduction could be attributed to the fact that our model represents an enhanced version of the IDoFew technique. Furthermore, in the Sentiment dataset, our model improved the F1-score by 4\% compared to the nearest baseline models, which is AraBERTv2 and 2\% against the IDoFew framework by utilising the same model and embedding. The 4\% difference between our model and AraBERTv2 resulted from targeting sentiment analysis task as a downstream task on the training of the AraBERTv2 model \cite{AraBERTv2-Exp-compMod-p1-antoun-etal-2020-arabert}. However, despite our model showing significant performance in the Sarcasm datasets, it did not match the efficiency of AceGPT LLM in these two datasets, achieving F1-scores of 78\% and 62\% in the sarcasm and Sa’7er datasets, respectively. It is worth noting that while AceGPT produced better results, it comes with high costs, whereas our model achieved significant results with minimal resource use. Moreover, our model raised the performance of one of the dialects dataset, achieving F1-scores of 67\% in the DART dataset, whereas Llama3 achieved 51\% in the dialects dataset. Additionally, in the MPOLD datasets, our models surpassed other models by achieving a 72\% F1-score.

To further investigate the availability of the FLICK framework for other LR languages beyond Arabic, we extended our experiments to two additional languages: Urdu (UOD and RUED) and Setswana. Figure~\ref{fig:low-resource-langs} demonstrates FLICK's capability to generalise across various LR languages. In the UOD dataset, we achieved an impressive accuracy of 89.6\% and an F1 score of 89.6\%. In contrast, the other Urdu dataset, RUED, yielded an accuracy of 41.1\% and an F1 score of 29.4\%. The performance disparity between these two Urdu datasets may stem from the inherent nature of the classification task, data points quantity, and the number of classes to predict. For the Setswana dataset, FLICK attained an accuracy of 50.3\% and an F1 score of 50.1\%.

These results not only demonstrate an enhancement in the performance of the SOTA model but also extend HR language techniques to include LR languages, thus promoting inclusivity and equality. Our framework has achieved remarkable results across three LR languages: 1) Arabic, 2) Urdu, and 3) Setswana, in addition to one HR language, English. Furthermore, our solution operates with a smaller number of parameters, approximately 200 million (0.2 billion), and outperforms models like Llama, which has 8 billion parameters, 40 times larger than AraBERTv2, and AceGPT, which has 13 billion parameters, 65 times larger than AraBERTv2, on most tasks.

Additionally, we have enhanced the SOTA framework IDoFew to improve the fine-tuned model for FL learning by refining the error correction mechanism in the IDoFew methodology \cite{IDoFew}. This is achieved by introducing a selection method for PLs in the intermediate layer by a PL Refinement model. This method ranks PLs based on how well the model identifies their patterns using accuracy metrics, allowing us to select the Top-K PLs for further fine-tuning. Consequently, this improvement in framework complexity,  maintains performance and positively impacts costs compared to LLMs and SOTA frameworks like IDoFew, ultimately reducing overall energy consumption and promoting accessibility and sustainability.


\begin{center}
\scriptsize
\setlength{\tabcolsep}{1.5pt}
\renewcommand{\arraystretch}{0.62}

\begin{tabular*}{\linewidth}{@{\extracolsep{\fill}}%
  l                                     
  S[table-format=2.2]
  S[table-format=2.2]
  S[table-format=2.2]
  S[table-format=2.2]
  S[table-format=2.2]
  c}                                    
\toprule
\textbf{Our Model Setting/ Dataset} &
\textbf{ArBNTopic} & \textbf{Sentiment} & \textbf{Sarcasm} &
\textbf{Dialects} & \textbf{MPOLD} &
\textbf{Lvl}\\
\midrule
Train 75\% Test 25\% & 51.46 & 61.85 & 63.54 & 27.73 & 70.29 & P\\
Train 80\% Test 20\% & 49.81 & 59.89 & 56.66 & 28.29 & 65.39 & P\\
Train 85\% Test 15\% & {\bfseries 53.58} & 61.78 & 61.15 & 28.01 & 65.64 & P\\
Train 90\% Test 10\% & 52.79 & 54.96 & 61.72 & 27.75 & 67.67 & P\\
Train 20\% Test 80\% & 47.86 & 62.60 & 66.41 & 29.08 & 61.33 & P\\
Train 15\% Test 85\% & 45.57 & 60.44 & 66.35 & 28.12 & 67.31 & P\\
Train 10\% Test 90\% & 39.43 & 61.22 & 60.63 & 28.18 & 67.66 & P\\
Top-K (Pseudo labels) = 10        & 34.75 & 62.57 & 50.45 & 26.68 & 67.60 & P\\
Top-K (Pseudo labels) = 15        & 42.64 & 60.61 & 60.98 & 27.43 & 66.10 & P\\
Top-K (Pseudo labels) = 25        & 51.23 & 61.27 & 60.57 & 28.84 & 66.40 & P\\
Top-K (Pseudo labels) = 30        & 48.30 & {\bfseries 63.30} & 63.48 & 28.28 & 67.09 & P\\
Top-K (Pseudo labels) = 40        & 48.57 & 60.38 & 55.61 & 28.35 & {\bfseries 71.57} & P\\
8-shots (labeled sample)    & 15.29 & 33.95 & 44.88 & 22.04 & 47.57 & P\\
16-shots(labeled sample)    & 22.42 & 49.57 & 50.83 & 20.27 & 45.45 & P\\
32-shots (labeled sample)   & 30.15 & 50.62 & 45.55 & 20.87 & 55.92 & P\\
\midrule
Gaussian Mixture Models     & 50.22 & 62.64 & 55.95 & 26.37 & 69.22 & C\\
Agglomerative Clustering    & 47.79 & 61.05 & 62.63 & 26.99 & 70.49 & C\\
Spectral Clustering         & 35.33 & 62.39 & {\bfseries 68.84} & 25.47 & 64.69 & C\\
\midrule
Sib - Top-K            & 44.97 & 63.08 & 62.25 & 25.79 & 60.70 & A\\
Sib - Top-K - Kmeans & 47.96 & 59.76 & 62.79 & 26.40 & 61.19 & A\\
Sib - Kmeans - Top-K   & 45.42 & 59.80 & 59.40 & {\bfseries 29.51} & 61.35 & A\\
\bottomrule
  \end{tabular*}
    \captionof{table}{F1 comparison of our model under three levels of change on five Arabic datasets.
The ``Lvl'' column marks the varied component: P – parameter settings, C – cluster-method choice, A – architecture order of sequential Information Bottleneck (SIB), k-means and Top-K.
The table serves to analyse how each modification affects performance.}
    \label{table-parameter-analysis1}
\end{center}

\subsection{Parameter Analysis}
In Table~\ref{table-parameter-analysis1}, we have performed different experiments by altering parameters, clustering methods, and architecture. The results show that performance generally declines with these variations, except in three of the 17 cases. The improvements in these instances could be due to the larger data volume from increased training set size and cluster numbers. The ArBNTopic dataset's significant improvement may relate to its encompassing 13 classes. We also evaluated clustering methods by replacing k-means with three different clustering techniques. The results showed a decrease in the F1-score for all methods, affecting the model's overall efficacy. An exception was noted with spectral clustering on the Sarcasm dataset. We also explored various structural changes within our framework. These modifications did not consistently enhance performance, thus diminishing the overall efficacy of our approach.

Our experiments demonstrate that selecting different Top-K values in the PL Refinement model, such as 10, 15, 25, and 30, yields varying effects on the PL-FT model's accuracy. Refer to Table~\ref{table-parameter-analysis1} for details. At this stage, our objective is to achieve a better initialisation for the FL-FT model, which will allow us to fine-tune only a few gold labels. We found that a Top-K value of 15 provides an experimentally effective initialisation for the FL-FT model, positively impacting overall performance. This may be attributed to the weights being in an optimal state for the FL-FT model to build upon, avoiding extremes that could require more computational resources, potentially compromising its generalisation and performance in the long run, especially in scenarios with limited data.

\begin{table}[H]
\scriptsize
\centering
\begin{tabular}{cccc}
\multicolumn{1}{l}{} & \multicolumn{3}{c}{}      
\\ \toprule
\multicolumn{1}{l}{\textbf{Datasets\textbackslash{}Metrics}} &
  \multicolumn{1}{l}{\textbf{Precision}} &
  \multicolumn{1}{l}{\textbf{Sensitivity}} &
  \multicolumn{1}{l}{\textbf{Specificity}} \\ \midrule
\textbf{ArBNTopic}   & 57.52\%    & 54.86\%    & 97.17\% \\ \midrule
\textbf{Sentiment}   & 62.96\%    & 60.18\%   & 81.17\% \\ \midrule
\textbf{Sarcasm}     & 70.61\%   & 60.07\%    & 60.07\% \\ \midrule
\textbf{Sa’7r}       & 63.71\%    & 62.84\%    & 62.84\% \\ \hline
\textbf{DART}        & 63.69\%   & 72.92\%   & 93.85\%
      \\ \midrule
\textbf{Dialects}    & 30.33\%    & 30.32\%    & 88.91\% \\ \midrule
\textbf{MPOLD}       & 80.23\%  & 69.30\%  & 69.30\% \\ 
\bottomrule
\end{tabular}
\caption{Error results of the FLICK model}
\label{exten-error-analysis}
\end{table}

Table~\ref{exten-error-analysis} presents the error results of the FLICK model across seven distinct datasets. It quantitatively evaluates FLICK's performance using three key metrics: Precision, Sensitivity (Recall), and Specificity. Each row details the percentage scores for these metrics on a specific dataset, such as ArBNTopic, Sentiment, and MPOLD. This allows for a comprehensive understanding of FLICK's classification accuracy, its ability to identify true positives, and its capability to correctly identify true negatives.

The table provides a granular evaluation of the FLICK model's performance beyond a single aggregated score. It allows for a nuanced understanding of the model's strengths and weaknesses across diverse text classification tasks and LR Arabic language. For instance, high specificity indicates FLICK's strong ability to correctly identify negative cases, which is crucial in applications where false positives are costly. Conversely, analysing sensitivity alongside precision helps pinpoint datasets where the model might struggle with either identifying all positive instances or minimising false alarms, guiding future improvements for robust performance in challenging linguistic environments.

\begin{figure}[ht!]
    \centering
    \scriptsize

    \includegraphics[width=0.50781\textwidth]{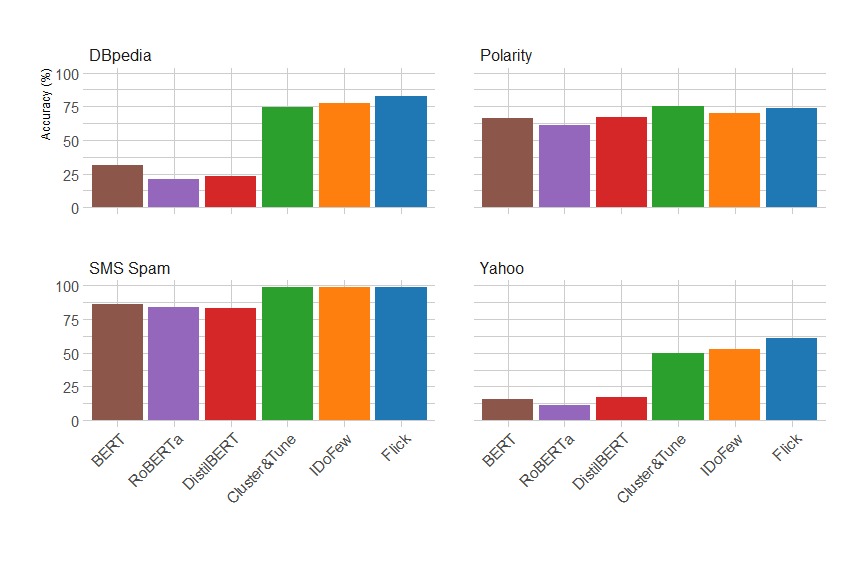}
    \centering
    \caption{Comparative analysis of the FLICK model alongside baseline and state-of-the-art models across various English datasets (SMS Spam, Polarity, Yahoo, and DBpedia). The analysis is structured into four individual bar charts, each dedicated to a specific dataset.}
    \label{fig:English-Datasets}
\end{figure}

\section{Case Study}
In this section, we ask the question `Is our framework applicable in widely used languages such as English?' Although our model was designed primarily for Arabic audiences with limited digital data resources, we have also conducted experiments to determine how effectively FLICK operates in prevalent languages such as English.

We have found that FLICK outperforms comparative models in HR language within the Yahoo Answers and DBpedia datasets, which is quite remarkable. This strongly demonstrates our model's scalability and adaptability, especially in tasks involving multiple languages. Figure \ref{fig:English-Datasets} illustrates our strong performance on English datasets. Our model possesses a more compact architecture than recent SOTA frameworks such as Cluster \& Tune \cite{shnarch-etal-2022-cluster} and IDoFew \cite{IDoFew}, which means it has fewer parameters and requires a less extensive fine-tuning process. This results in reduced computational time and resources, simplifying the complexity found in previous frameworks.

\section{Conclusion}
We have developed a novel FLICK framework to address the issue of inclusive minorities of LR languages and linguistic inequality caused by limited labelled web data via intermediate label learning with a particular focus on Arabic datasets. FLICK employs selective PLs to mitigate noise, focusing on choosing the most pertinent examples and reducing used models. Our extensive experiments have shown a marked enhancement in performance, scalability, accessibility, and energy cost across datasets that feature a dearth of labelled data, mainly in Arabic and including Urdu and Setswana texts. Moving forward, we will extend our experiments to additional languages, including Pashto and more Bantu languages such as Swahili.

Ethical considerations in FLICK primarily revolve around the potential for bias propagation: PLs and clustering, while effective for data augmentation, can inadvertently amplify existing biases present in the limited labelled data, potentially leading to unfair or discriminatory outcomes when applied in sensitive domains. Therefore, careful auditing of the initial labelled dataset and the resulting PLs is crucial to mitigate unintended societal impacts.

\section*{Limitations}
A significant challenge lies in optimally selecting the number of clusters, as non-parametric methods can be computationally intensive. Furthermore, the quality and quantity of available Arabic datasets critically affect model performance, potentially introducing biases or impeding the learning of complex patterns and generalisation to new domains. Finally, the inherent opaqueness of deep learning models and the risk of propagating training data biases necessitate careful consideration of fairness and responsible application to prevent discriminatory practices.

\bibliography{custom}

\end{document}